\pdfoutput=1
\documentclass[sigconf]{acmart}
\usepackage{epsfig}
\usepackage{graphicx}
\usepackage{amsmath}

\usepackage{amssymb}
\usepackage{bm}
\usepackage{enumitem}
\usepackage{multirow}
\usepackage{caption}
\usepackage{soul}

\DeclareMathOperator{\argmin}{argmin} 

\newcommand{\etal}{\textit{et al. }}
\newcommand{\eg}{\textit{e.g., }}
\newcommand{\ie}{\textit{i.e., }}

\AtBeginDocument{%
  \providecommand\BibTeX{{%
    \normalfont B\kern-0.5em{\scshape i\kern-0.25em b}\kern-0.8em\TeX}}}

\setcopyright{acmcopyright}
\copyrightyear{2020}
\acmYear{2020}

\copyrightyear{2020}
\acmYear{2020}
\setcopyright{acmcopyright}
\acmConference[MM '20]{Proceedings of the 28th ACM International Conference on Multimedia}{October 12--16, 2020}{Seattle, WA, USA}
\acmBooktitle{Proceedings of the 28th ACM International Conference on Multimedia (MM '20), October 12--16, 2020, Seattle, WA, USA}
\acmPrice{15.00}
\acmDOI{10.1145/3394171.3413555}
\acmISBN{978-1-4503-7988-5/20/10}
\begin{document}
\fancyhead{}
\author{Zhenyu Wu}
\authornote{The first two authors Wu and Hoang contributed equally to this research.}
\email{wuzhenyu_sjtu@tamu.edu}
\affiliation{%
  \institution{Texas A\&M University}
}

\author{Duc Hoang}
\email{hoangd@tamu.edu}
\authornotemark[1]
\affiliation{%
  \institution{Texas A\&M University}
}

\author{Shih-Yao Lin}
\email{shihyaolin@tencent.com}
\affiliation{%
  \institution{Tencent America}
}
\author{Yusheng Xie}
\authornote{Work done prior to Amazon.}
\email{yushx@amazon.com}
\affiliation{%
  \institution{Amazon Web Services}
}

\author{Liangjian Chen}
\email{liangjc2@ics.uci.edu}
\affiliation{%
  \institution{University of California, Irvine}
}

\author{Yen-Yu Lin}
\email{lin@cs.nctu.edu.tw}
\affiliation{%
  \institution{National Chiao Tung University}
}

\author{Zhangyang Wang}
\email{atlaswang@utexas.edu}
\affiliation{%
  \institution{University of Texas at Austin}
}

\author{Wei Fan}
\email{davidwfan@tencent.com}
\affiliation{%
  \institution{Tencent America}
}

\begin{abstract}
Estimating the 3D hand pose from a monocular RGB image is important but challenging.
A solution is training on large-scale RGB hand images with accurate 3D hand keypoint annotations. 
However, it is too expensive in practice. 
Instead, we have developed a learning-based approach to synthesize realistic, diverse, and 3D pose-preserving hand images under the guidance of 3D pose information. 
We propose a 3D-aware multi-modal guided hand generative network (\textbf{MM-Hand}), together with a novel geometry-based curriculum learning strategy. 
Our extensive experimental results demonstrate that the 3D-annotated images generated by {\em MM-Hand} qualitatively and quantitatively outperform existing options. 
Moreover, the augmented data can consistently improve the quantitative performance of the state-of-the-art 3D hand pose estimators on two benchmark datasets. 
The code will be available at \href{https://github.com/ScottHoang/mm-hand}{\color{blue}{\textit{https://github.com/ScottHoang/mm-hand}}}.
\end{abstract}

\title{MM-Hand: 3D-Aware Multi-Modal Guided Hand Generative Network for 3D Hand Pose Synthesis}



\keywords{3D Hand-Pose, Multi-Modal, Curriculum Learning, Conditional Generative Adversarial Nets}

\maketitle

\section{Introduction}

3D hand pose estimation is an important and active research topic due to its versatile applications in sign language recognition~\cite{xiao2020skeleton}, HCI (human-computer interaction)~\cite{lin2017design,lin2012action,lin2013airtouch}, healthcare and entertainment~\cite{krejov2016real, bhuyan2011hand, malima2006fast}. 
Some HCI applications such as typing\footnote{https://uploadvr.com/frl-pinchtype-ar-vr-keyboard/} highly rely on accurately estimated hand poses.
Conventional HCI applications use depth sensors to capture hand information and infer hand poses~\cite{tompson2014real, sun2015cascaded, lin2010real, hung2016re,lin2017learning}.
In the past few years, there has been growing interest in HCI applications where pose estimation or tracking from single RGB images is applied, usually to take advantage of the ubiquity of RGB cameras. 
For example, the exciting dancing application~\cite{Chan_2019_ICCV} and the popular personalized media creation app, Humen.ai\footnote{https://www.humen.ai}, both require RGB-based body-hand pose estimation and tracking. 
However, without depth information, estimating the 3D hand pose from a monocular RGB image is an ill-posed problem.
Recent studies~\cite{baek2019pushing,boukhayma20193d,cai2018weakly,ge20193d,yang2019disentangling,zimmermann2017learning} resort to a large number of training images with the corresponding 3D hand pose annotations. 
Nevertheless, acquiring large-scale hand datasets with 3D annotations is labor-intensive and very expensive. 
Compared with 2D annotations, 3D annotations of real RGB hand images are much more difficult to be reliably labeled by humans.

A promising alternative emerges from synthesizing hand images. 
With 3D computer graphics software (\eg Blender and Maya) and chroma key compositing, synthetic hand data can be generated with accurate 3D annotations~\cite{mueller2017real,zimmermann2017learning}. 
In this way, a large hand image dataset with various poses, skin textures, shapes, background, lighting conditions, and object interactions can be systematically synthesized. 
However, synthesizing hand images to train 3D hand pose estimators is limited in two aspects. 
Firstly, estimators trained on the synthetic data often fail to generalize due to the visual domain gap between non-photo-realistic synthetic data and real images.
Such a domain gap has been widely documented and investigated in prior work~\cite{shrivastava2017learning,chentagan,mueller2018ganerated,chen2020automated}.
Secondly, building hand models with various textures or shapes requires laborious 3D geometric modeling and rendering, which are relatively less studied.

Recent work on the generation of pose-guided person image ~\cite{esser2018variational,liu2019liquid,ma2017pose,ma2018disentangled,song2019unsupervised,zhu2019progressive, neverova2018dense,si2018multistage,pumarola2018unsupervised} has made significant progress by human body pose transfer, \eg swapping the pose of a person image into a target pose while maintaining other visual appearance details. 
Those generated realistic person images are blended into the training process to improve the person re-ID task. 
Inspired by those, our goal is to generate realistic and diverse hand images with accurate pose annotations.
However, there are two challenges specific to the hand domain~\cite{li2019survey}:
\textit{occlusion} (\ie, various 3D hand movements will always make some finger parts invisible from 2D images) and \textit{self-similarity} (\ie the five fingers of the same hand share similar appearance and structure, making them indistinguishable).
This paper makes the following three-fold contributions:
\begin{itemize}[leftmargin=*]
\item Our proposed framework, {\em 3D-Aware Multi-modal Guided Hand Generative Network} (\textbf{MM-Hand}), carries out the first attempts to generate hand images under the guidance of 3D poses. The proposed {\em MM-Hand} is able to improve the realism, increase the diversity, and preserve the 3D pose of the generated images simultaneously.
\item {\em MM-Hand} is trained with a novel geometry-based curriculum learning strategy. Starting with easy pose-images pairs, we gradually increase the training task difficulty.
\item Extensive experiments demonstrate that our generated hand images can consistently improve 3D hand pose estimation, across two strong pose estimators and two hand pose datasets.
\end{itemize}
\section{Related Work}
\subsection{Data Augmentation for Hand Pose Estimation}
Generative adversarial networks (GANs) have demonstrated strong promise in synthesizing training data~\cite{zhang2019dada,li2019triple,chen2020dggan}. Shrivastava \etal ~\cite{shrivastava2017learning} proposes {\em SimGAN}, which improves the realism of a simulator's rendered data by using unlabeled real data while preserving the annotation information from the simulator. The processed data by {\em SimGAN} are then leveraged to train a hand pose estimator. Mueller \etal~\cite{mueller2018ganerated} presents {\em GeoConGAN}, whose generated images preserves their hand pose by a geometric consistency loss. These data augmentation approaches focus on image translation from the synthetic hands to real hands (based on an existing synthetic simulator). In contrast, we directly generate realistic hand images from 3D poses and synthetic depth maps, which is more challenging.

Zimmermann and Brox~\cite{zimmermann2019freihand} introduce the first large-scale, multi-view hand image dataset, which includes both 3D hand pose and shape annotations. 
The annotation is achieved by an iterative, semi-automated ``human-in-the-loop'' approach, which includes hand fitting optimization to infer the 3D pose and shape for each sample.

\subsection{Pose Guided Person Image Generation}

Isola \etal~\cite{isola2017image} proposes {\em Pix2Pix} to translate sketch to synthesize photos from label maps, reconstructing objects from edge maps and colorizing images, using paired data.
Zhu \etal~\cite{zhu2017unpaired} present the {\em CycleGAN} to work under the unpaired settings by introducing a cycle consistency loss between source and target domains. Since then, (unpaired) image translation has been popular in various applications including image enhancement~\cite{jiang2019enlightengan,kupyn2019deblurgan,m2019all}, style transfer~\cite{yang2019controllable},  interactive image editing~\cite{yang2020deep}, and domain adaptation~\cite{hoffman2018cycada,murez2018image,wu2019delving,wu2018towards,wang2019privacy}.

As the first work focusing on pose-guided human image generation, Ma \etal~\cite{ma2017pose} proposes a two-stage network {\em $\text{PG}^2$}, where a person image under the target pose is first coarsely generated and then refined. Ma \etal~\cite{ma2018disentangled} then improves the control over image generation by disentangling and separately encoding the three modes of variation (foreground, background, and pose information) into embedded features.
Esser \etal~\cite{esser2018variational} combine {\em Variational Auto-Encoder (VAE)}~\cite{kingma2013auto} and {\em U-Net}~\cite{ronneberger2015u} to disentangle appearance and poses. Their work presents a {\em U-Net} that maps from shape to target image, conditioned on a {\em VAE} latent representation for preserving appearances.

Si \etal~\cite{si2018multistage} adopts multistage adversarial losses separately for the foreground and background generation, which fully exploits the multi-modal characteristics of generative loss to generate more realistic looking images.
Neverova \etal~\cite{neverova2018dense} proposes a combination of surface-based pose estimation and deep generative models to perform accurate pose transfer.
Siarohin \etal~\cite{siarohin2018deformable} introduces deformable skip connections in the generator to deal with pixel-to-pixel misalignments caused by the pose differences, together with a nearest-neighbor loss.
Pumarola \etal~\cite{pumarola2018unsupervised} proposes a pose conditioned bidirectional generator that maps back the initially rendered image to the original pose, hence being directly comparable to the input image without any training image.
Li \etal~\cite{li2019dense} proposes to estimate dense and intrinsic 3D appearance flow to guide the transfer of pixels between poses better. They generate 3D flow by fitting a 3D model to the given pose pair and project them back to the 2D plane to compute the dense appearance flow for training.
Song \etal~\cite{song2019unsupervised} addresses unsupervised person image generation by decomposing it into semantic parsing transformation and appearance generation.
Zhu \etal~\cite{zhu2019progressive} proposes a progressive pose attention transfer network composed of a cascaded {\em Pose-Attentional Transfer Blocks (PATBs)}.
Liu \etal~\cite{liu2019liquid} tackles the human motion imitation, appearance transfer, and novel view synthesis within a unified framework. 
Unlike pose guided person images generation, pose guided hand generation can be much more subtle and difficult, due to the inherently strong self-similarity and the self-occlusion.

\subsection{3D Hand Pose Estimation from a Single Image}
Zimmermann and Brox~\cite{zimmermann2017learning} proposes the first learning-based approach to estimate the 3D hand pose from a single RGB image.
Their approach consists of three building blocks: {\em HangSegNet} for obtaining hand mask by segmentation, {\em PoseNet} for localizing a set of hand keypoints in score maps, and {\em PosePrior} network for estimating 3D structure conditioned on the score maps).
Cai \etal~\cite{cai2018weakly} proposes a weakly-supervised method to generate depth maps from predicted 3D poses, which then serves as weak supervision for 3D pose regression.
Chen \etal~\cite{chen2020dggan} presents a {\em Depth-image Guided GAN (DGGAN)} to generate realistic depth maps conditioned on the input RGB image and use the synthesized depth maps to regularize the 3D hand pose estimators.
Some studies~\cite{boukhayma20193d, ge20193d,baek2019pushing} tackle the challenging task of 3D hand shape {\it and} pose estimation from a single RGB image.
Boukhayma \etal~\cite{boukhayma20193d} combines a deep convolutional encoder with a generative hand model as the decoder. Ge \etal~\cite{ge20193d} proposes a graph CNN-based method to reconstruct 3D mesh of hand surface that contains rich information of both the 3D hand shape and pose. Baek \etal~\cite{baek2019pushing} adopts a compact parametric 3D hand model that represents deformable and articulated hand meshes. Yang \etal~\cite{yang2019disentangling} proposes a disentangled VAE that allows for sampling and inference of variation factors, \eg content, poses, and camera views.
\section{Our Approach}
\begin{figure*}
\centering{
  \includegraphics[width=1.00\textwidth]{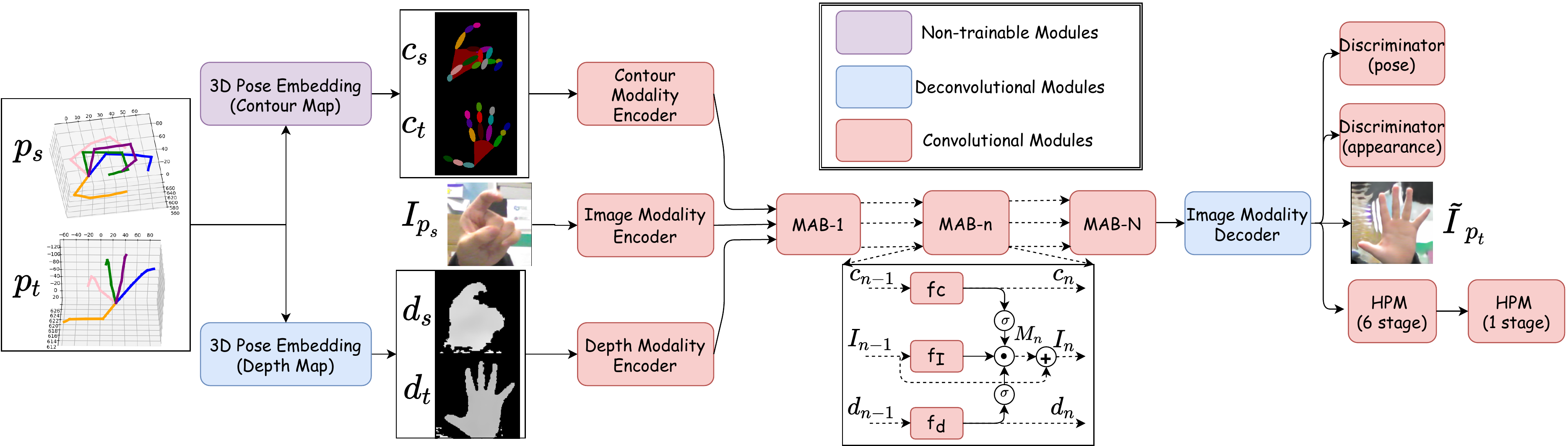}
  \caption{An overview of our proposed {\em MM-Hand} Model. {\em MM-Hand} mainly consists of four modules: \textit{3D Pose Embedding}, \textit{Multi-modality Encoding}, \textit{Progressive Transfer} and \textit{Image Modality Decoding}.
}
  \label{MM-Hand}
    \vspace{-1em}
}
\end{figure*}

\subsection{Problem Formulation}
\paragraph{\textbf{Goal}} 
Given a target 3D hand pose $\bm{p}_t$, and a source image $\bm{I}_{p_s}$ under a source 3D pose $\bm{p}_s$, our goal is to generate a new image $\bm{\Tilde{I}}_{p_t}$ following the appearance of $\bm{I}_{p_s}$, under the guidance of $\bm{p}_s$ and $\bm{p}_t$. The generation can be formulated as: 
\begin{align}
    <\bm{I}_{p_s}, \bm{p}_s, \bm{p}_t> \rightarrow \bm{\Tilde{I}}_{p_t}
\end{align}

\paragraph{\textbf{Evaluation Protocols}} 
The generated hand image $\bm{\Tilde{I}}_{p_t}$ is expected to resemble the ground truth hand image $\bm{I}_{p_t}$ in both visual realism and pose consistency. The visual realism is evaluated by natural image quality metrics (\eg SSIM and IS). 
The pose consistency is approximated by pose joints alignment (\eg PCKb).
Considering both visual realism and pose consistency, we further evaluate the quality of generated hand images on the visual task of 3D hand pose estimation.
Details of the evaluation metrics are given in section 4.1.

\paragraph{\textbf{Pose Representations}} 
In this paper, we use the $21$-joint hand model ($K=21$). 
3D poses are represented as $3 \times K$ matrices: $[J_1^{xyz}, \dots, J_K^{xyz}] \in \mathcal{R}^{3\times K} (K=21)$.  
2D poses are represented as $K$ probability heat maps: $\{\bm{H}_i\}_{i=1}^K \in \mathcal{R}^{K \times H \times W} (K=21)$, where $\bm{H}_i$ is a Gaussian distribution centered at the location of the $i_{th}$ joint.
\subsection{Architecture Overview}
As is shown in Figure \ref{MM-Hand}, the proposed \textit{3D-Aware Multi-modal Guided Hand Generative Network} (MM-Hand) is composed of 4 modules: \textit{3D pose embedding}, \textit{multi-modality encoding}, \textit{progressive transfer} and \textit{image modality decoding}.
\begin{figure}
\centering{
  \includegraphics[width=0.485\textwidth]{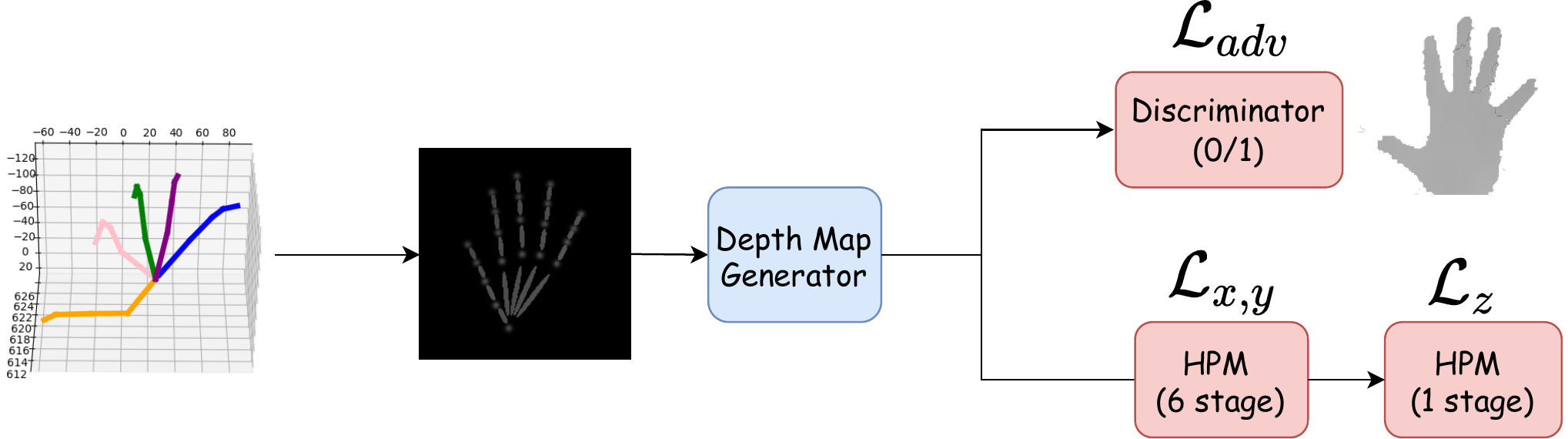}
  \caption{A detailed look into the depth map generation model for 3D pose embedding}.
  \vspace{-1em}
  \label{fig:depth-map}
}
\end{figure}

\subsubsection{3D Pose Embedding}
\paragraph{\underline{Contour Map Embedding}}
Given the camera's intrinsic matrix $\bm{K}$ and extrinsic matrix $[\bm{R} \mid \bm{-RC}]$, we obtain the projection matrix $\bm{P}=\bm{K}[\bm{R} \mid \bm{-RC}]$ which transforms homogeneous 3D world coordinates to homogeneous 2D image coordinates. Firstly, we represent the $K$ joints with 2D coordinates with a sparse pose map, using erosion and dilation. Secondly, we connect the keypoints on finger with solid ellipsoid using different colors. Lastly, A palm surrograte is formed by connecting a polygon from the base of each finger and the wrist. The contours map of $\bm{c}_{p_s}$ and $\bm{c}_{p_t}$ are generated as the embeddings of the 3D pose $\bm{p}_s$ and $\bm{p}_t$.

\paragraph{\underline{Depth Map Embedding}}
There are datasets that contain depth maps paired with annotated 3D hand poses, \eg ICVL ~\cite{tang2014latent}, NYU ~\cite{tompson2014real} and MSRA ~\cite{sun2015cascaded}. The annotated hand poses in ICVL, MSRA and NYU have $16$ joints, $21$ joints, and $14$ joints. We choose the MSRA dataset annotated with $21$ joints to be consistent with those benchmark datasets.

With the help of external datasets with paired depth maps and 3D hand poses, we can learn a depth map generator which convert a 3D hand pose to a depth map. In Figure \ref{fig:depth-map}, the depth map generator takes the raw 3D hand pose ($\bm{p}_s$ and $\bm{p}_t$) as input and outputs depth maps ($\bm{d}_{p_s}$ and $\bm{d}_{p_t}$). 

The depth map generator is a Pix2Pix model regulated by a discriminator and a pair of 2D/3D key-points estimators. The key-points estimators are 6-stage HPM~\cite{wei2016convolutional}, and 1-stage HPM~\cite{chen2020dggan}. Both are pre-trained on the MSRA's pose estimation task. Note that we first project the 3D hand pose onto a 2D image before feeding it to the generator.

\subsubsection{Multi-Modality Encoding}
The modality encoder for the image modality, the contour map modality and the depth map modality are consistently adopting two convolution layers. Before encoding, we concatenate $\bm{c}_{p_s}$ with $\bm{c}_{p_t}$, and $\bm{d}_{p_s}$ with $\bm{d}_{p_t}$. Specifically,
\begin{align}
    \bm{c}_0 = f_c^e(\bm{c}_{p_s} \mathbin\Vert \bm{c}_{p_t}), \bm{d}_0 = f_d^e(\bm{d}_{p_s} \mathbin\Vert \bm{d}_{p_t})\ \text{and}\ \bm{I}_0=f_I^e(\bm{I}_{p_s}),
\end{align}
where $f_c^e$, $f_d^e$ and $f_I^e$ are the modality encoders for the contour maps, the depth maps and the images respectively.

\subsubsection{Progressive Transfer}
Our \textit{progressive transfer} module inherits the ResNet Generator proposed in ~\cite{johnson2016perceptual, isola2017image}. It consists of multiple cascaded Multi-stream Attentional Blocks (MABs) (shown in Figure \ref{MM-Hand}), variants of the ResNet blocks in ~\cite{johnson2016perceptual, isola2017image}. MAB is similar to the Pose Attention Transfer Block (PATB) proposed in ~\cite{zhu2019progressive}. Starting from the initial image modality $\bm{I}_0$, the contour map modality $\bm{c}_0$ and depth modality $\bm{d}_0$, {\em MM-Hand} progressively updates these three modalities through a sequence of MABs. Then, deconvolution is used to decode the output image modality $\bm{I}_N$ to generate $\bm{\Tilde{I}}_{p_t}$. The final contour map modality $\bm{c}_N$ and depth map modality $\bm{d}_N$ are discarded after inference.

All MABs share an identical structure. 
%
One MAB's output becomes the input for the next MAB block.
For example, 
the input of the $n\text{-th}$ block consists of $\bm{I}_{n-1}$, $\bm{c}_{n-1}$ and $\bm{d}_{n-1}$. The specific modules within each MAB are described in the following paragraphs.

\paragraph{\underline{Attention Masks}} 
Inspired by Zhu \etal~\cite{zhu2019progressive}, the attention masks $\bm{M}_n$, whose values are between 0 and 1, indicate the importance of every element in the image modality. $\bm{M}_n$ are computed from the contour modality $\bm{c}_{n-1}$ and the depth modality $\bm{d}_{n-1}$. The contour modality $\bm{c}_{n-1}$ incorporates both the source contour map $\bm{c}_{p_s}$ and the target contour map $\bm{c}_{p_t}$. Likewise, the depth modality $\bm{d}_{n-1}$ incorporates both the source depth map $\bm{d}_{p_s}$ and the target depth map $\bm{d}_{p_t}$. The attention mask is computed as element-wise product of $\sigma(f_c(\bm{c}_{n-1}))$ and $\sigma(f_d(\bm{d}_{n-1}))$, where $\sigma$ is an element-wise sigmoid function and $f_c$ and $f_d$ are ResNet Blocks. Specifically, the attention masks $\bm{M}_n$ are obtained via:
\begin{align}
    \bm{M}_n=\sigma(f_c(\bm{c}_{n-1})) \odot \sigma(f_d(\bm{d}_{n-1})).
\end{align}

\paragraph{\underline{Image Modality Update}} By multiplying the transformed image codes with the attention mask $\bm{M}_n$, image code $\bm{I}_n$ at certain locations are either preserved or suppressed. The output of the element-wise product is added by $\bm{I}_{n-1}$, via a residual connection. The residual connection helps preserve the original image modality. $f_I$ is again a ResNet Block. The image modality $\bm{I}_n$ is updated by: 
\begin{align}
    \bm{I}_n=\bm{M}_n \odot f_I(\bm{I}_{n-1})+\bm{I}_{n-1}.
\end{align} 

\paragraph{\underline{Discriminators}}
We adopt the same two discriminators in Zhu \etal ~\cite{zhu2019progressive}: an appearance discriminator and a pose discriminator. They are denoted as $D_a$ and $D_p$, respectively. $D_a(\bm{I}_{p_s}, \bm{\Tilde{I}}_{p_t})$ describes how well $\bm{\Tilde{I}}_{p_t}$ resembles the source image $\bm{I}_{p_s}$ in appearance. $D_p(\bm{p}_t, \bm{\Tilde{I}}_{p_t})$ shows how well $\bm{\Tilde{I}}_{p_t}$ is aligned with the target pose $\bm{p}_t$.

\subsubsection{Image Modality Decoding}
We take the output image modality $\bm{I}_N$ from the $N\text{-th}$ MAB, and generate $\bm{\Tilde{I}}_{p_t}$ from $\bm{I}_N$ via the image modality decoder.

\subsection{Training}
\subsubsection{Geometry-based Curriculum Training and Inference with Nearest Neighbor Match}
Given two 3D hand poses $\bm{u}$ and $\bm{v}$, we define the pose distance between $\bm{u}$ and $\bm{v}$ as:
\begin{align} \label{eq:pose_distance}
    d(\bm{u}, \bm{v}) = \frac{1}{\pi}cos^{-1}(\frac{<f(\bm{u}), f(\bm{v})>}{\left\lVert f(\bm{u}) \right\rVert \left\lVert f(\bm{v}) \right\rVert})
\end{align}
where $f(\cdot)$ describes the ``identity'' of a hand pose. Each hand pose is expressed as the concatenation vector of its $21$ 3D keypoints, \ie $\bm{u}=[u_x^1,u_y^1,u_z^1,\dots,u_x^{21},u_y^{21},u_z^{21}] \in \mathbb{R}^{63}$. The ``identity'' $f(\bm{u})$ of $\bm{u}$ is defined as a vector of 3 components: $d_{tips}(\bm{u})$, $d_c(\bm{u})$ and $\sqrt{\mathcal{A}(\bm{u})}$.
\begin{align}
    f(\bm{u})=[d_{tips}(\bm{u}), d_c(\bm{u}), \sqrt{\mathcal{A}(\bm{u})}]
\end{align}
$d_{tips}(\bm{u})$ is a collection of pair-wise euclidean distance between the tip keypoint (thumb, index, mid, ring and pinky) and the palm.
$d_c(\bm{u})$ is the distance between the centroid and the palm. $\sqrt{\mathcal{A}(\bm{u})}$ is the square root of the contour area of the convex hull formed by the $21$ 2D keypoints after projection.

\begin{figure}[!htb]
	\centering {
		\includegraphics[width=0.47\textwidth]{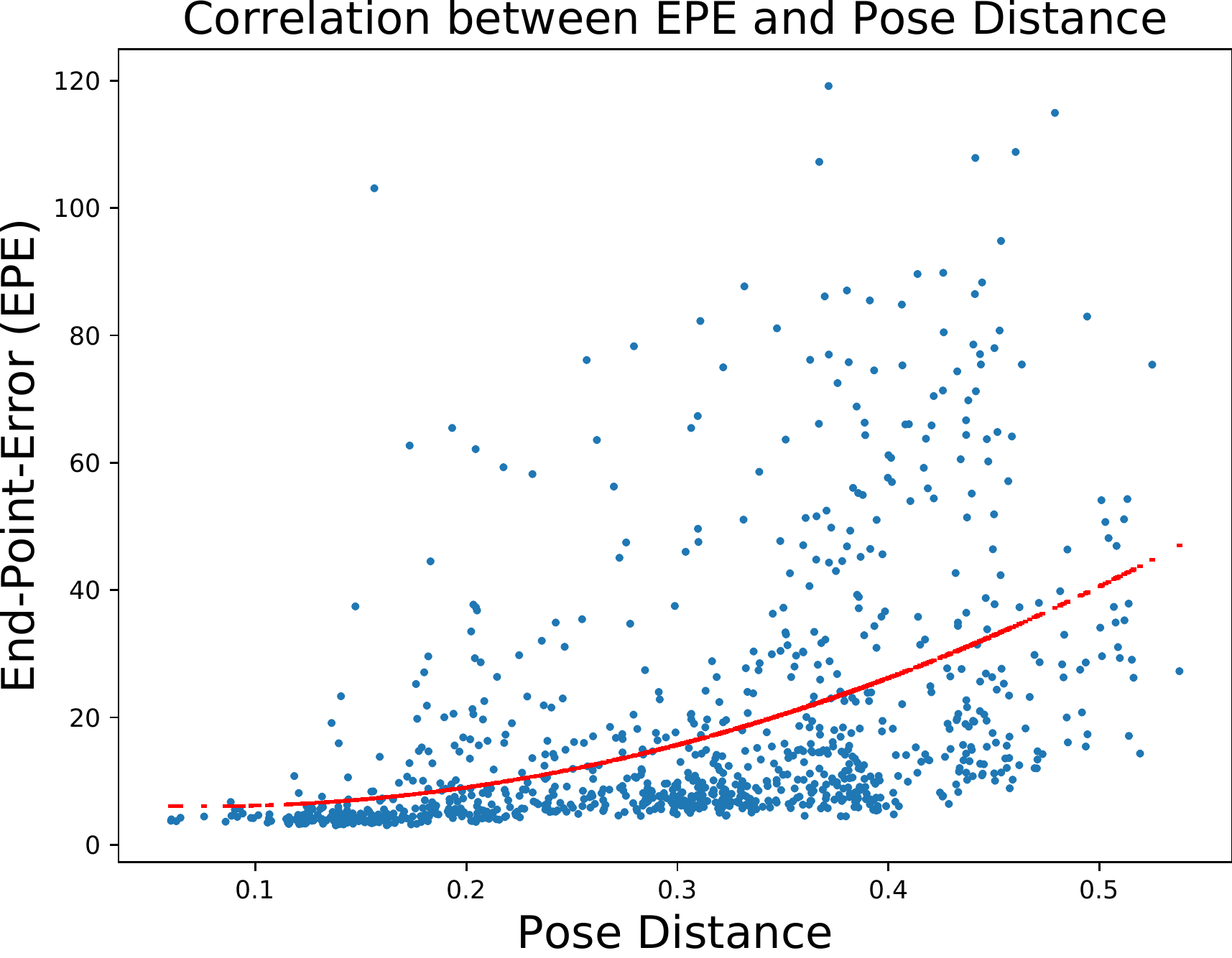}
	}
	\vspace{-2em}
	\caption{On STB, we first randomly select $1,000$ hand images $\bm{I}_{p_s}$s with their paired poses $\bm{p}_s$s from the testing set $\mathcal{X}_{te}$. Then we randomly select $1,000$ hand images $\bm{I}_{p_t}$s with their paired poses $\bm{p}_t$s from the training set $\mathcal{X}_{tr}$. We randomly make $1,000$ pairs of $(\bm{I}_{p_s}, \bm{p}_s)$ and $(\bm{I}_{p_t}, \bm{p}_t)$. We generate $1000$ $\bm{\Tilde{I}}_{p_t}$s from $<\bm{I}_{p_s}, \bm{p}_s, \bm{p}_t>$. We use a parabola function (the red curve) to fit the scattered points using the least square method. We empirically found that the End-Point-Error (EPE) between the $\bm{\Tilde{I}}_{p_t}$ and the ground-truth $\bm{I}_{p_t}$ is positively correlated with the pose distance (defined in Eq. (\ref{eq:pose_distance})) between $\bm{I}_s$ and $\bm{I}_t$.}
	\label{sbt_epe_dist_correlation}
	  \vspace{-1em}
\end{figure}

\noindent\paragraph{\underline{Geometry-based Curriculum Training (GCT)}} Based on the observation that EPE is positively correlated with the pose distance (shown in Figure \ref{sbt_epe_dist_correlation}), we hypothesize that the level of difficulty to generate target hand image $\bm{\Tilde{I}}_{p_t}$ from source hand image $\bm{I}_{p_s}$ is positively correlated with the 3D pose distance between $\bm{p}_s$ and $\bm{p}_t$. Hence in the training stage, we first randomly make pairs of hand images. Then we compute the 3D pose distance for each paired image. For each training epoch, {\em MM-Hand} is fed by the data loader with hand pairs progressively from the easiest (smallest pose distance) pair to the hardest (largest pose distance) pair.

\noindent\paragraph{\underline{Inference with Nearest Neighbor Match (INNM)}} In the inference stage, given a target 3D hand pose $\bm{p}_t$, we find the best matched source hand image $\bm{I}_{p_s}$ in the training hand images whose pose $\bm{p}_s$ is closest to $\bm{p}_t$ in pose distance.

\subsubsection{Loss Function}
The joint loss function is a nested sum of various types of loss functions. Specifically:
\begin{align}
\mathcal{L}_{joint} &= \alpha \mathcal{L}_{adv}+ \mathcal{L}_{app} + \mathcal{L}_{pose}\\
\mathcal{L}_{app} &= \tau_1 \mathcal{L}_1 + \tau_2 \mathcal{L}_{p}\\
\mathcal{L}_{pose} &= \gamma_1 \mathcal{L}_{x,y} + \gamma_2 \mathcal{L}_z,
\end{align}
where $\mathcal{L}_{adv}$ denotes the adversarial loss, $\mathcal{L}_{app}$ measures the appearance difference of the generated hand image and the target hand image, and $\mathcal{L}_{pose}$ is a 3D hand pose estimation task loss. 
$\alpha, \tau_{1,2}, \gamma_{1,2}$ represent the corresponding weights and they are determined empirically. The adversarial loss is defined as:
\begin{align}
& \mathcal{L}_{adv} = \mathbb{E}_{\bm{I}_{p_s},\bm{I}_{p_t},\bm{p}_t}\{\text{log}[D_a(\bm{I}_{p_s}, \bm{I}_{p_t}) \cdot D_p(\bm{p}_t, \bm{I}_{p_t})]\}, \nonumber \\
& + \mathbb{E}_{\bm{I}_{p_s},\bm{\Tilde{I}}_{p_t},\bm{p}_t}\{\text{log}[(1-D_a(\bm{I}_{p_s},\bm{\Tilde{I}}_{p_t})) \cdot (1-D_p(\bm{p}_t,\bm{\Tilde{I}}_{p_t}))]\},
\end{align}
where $\bm{\Tilde{I}}_{p_t}=G(\bm{I}_{p_s}, \bm{p}_s, \bm{p}_t)$. $G(\cdot)$ is the progressive transfer module.
$\mathcal{L}_1$ denotes the pixel-wise $\ell_1$ loss computed between the generated hand image and the target image, \ie $\mathcal{L}_1=\left\lVert \bm{\Tilde{I}}_{p_t} - \bm{I}_{p_t} \right\rVert_1$.
$\mathcal{L}_p$ is a perceptual loss ~\cite{johnson2016perceptual} widely used in style transfer and super resolution defined as:
\begin{align}
\mathcal{L}_p = \frac{1}{C_iH_iW_i} \left\lVert \phi_i(\bm{\Tilde{I}}_{p_t})-\phi_i(\bm{I}_{p_t})\right\rVert_2^2,
\end{align}
where $\phi_i$ is the $i_{th}$ layer of a pretrained VGG-16 network. We empirically use the $\textit{\text{conv}3\_3}$ layer.
$\mathcal{L}_{x,y}$ denotes the 2D hand pose estimation loss:
\begin{align}
    \mathcal{L}_{x,y} = \frac{1}{6K} \displaystyle \sum_{s=1}^{6} \displaystyle \sum_{i=1}^{K} \left\lVert \bm{H}_i^s - \bm{H}_i^* \right\rVert_F^2, 
\end{align}
where $\{\bm{H}_i^*\}_{i=1}^K(K=21)$ is the ground truth 2D poses in heat maps and $6$ is the number of stages in HPM. $\mathcal{L}_z$ denotes the depth estimation loss: 
\begin{align}
    \mathcal{L}_{z} =  \frac{1}{K} \sum_{i = 1}^K
    \left\{
    \begin{aligned}
       \frac{1}{2}(\bm{Z}_i - \bm{Z}_i^*)^2, |\bm{Z}_i - \bm{Z}_i^*| \le 1 \\
       |\bm{Z}_i - \bm{Z}_i^*|-0.5, otherwise,
    \end{aligned}
    \right.
\end{align}
where $\{\bm{Z}_i^*\}_{i=1}^K(K=21)$ is the ground truth relative depth.
\section{Experiments}

\begin{figure}[t]
    \begin{center}
	\begin{tabular}{llll}
		\includegraphics[width=0.112\textwidth]{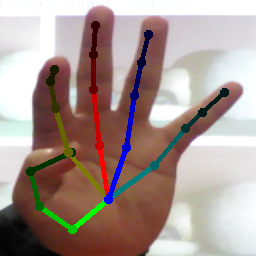}
		\includegraphics[width=0.112\textwidth]{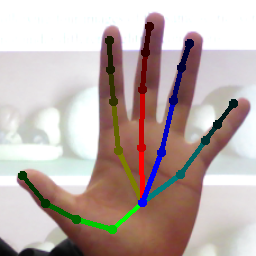}
		\includegraphics[width=0.112\textwidth]{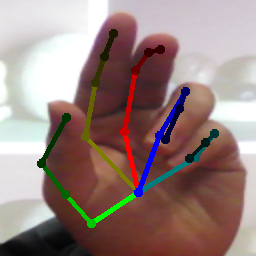}
		\includegraphics[width=0.112\textwidth]{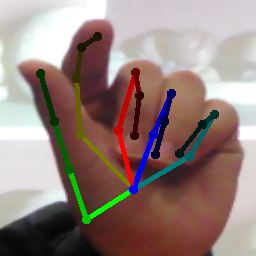}
		\label{fig:stb}
	\end{tabular}
	\end{center}
	
	\begin{center}
	\begin{tabular}{llll}
		\includegraphics[width=0.112\textwidth]{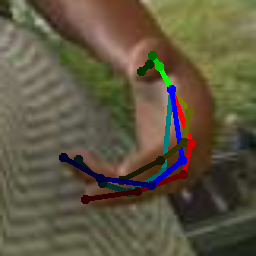}
		\includegraphics[width=0.112\textwidth]{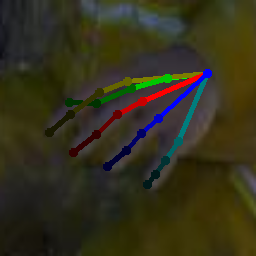}
		\includegraphics[width=0.112\textwidth]{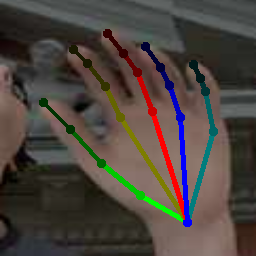}
		\includegraphics[width=0.112\textwidth]{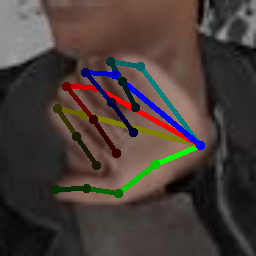}
	    \label{fig:rhp}
	\end{tabular}
	\end{center}
	\vspace{-1em}
	\caption{Some examples of the two benchmark hand datasets used for qualitative and quantitative evaluation. 
	\textbf{Top Row:} The STB dataset ~\cite{zhang20163d} contains real hand images with $3$D keypoints. 
	\textbf{Bottom Row:} The RHP dataset~\cite{zb2017hand} provides synthetic hand images with $3$D hand keypoint annotations.
	With more diverse poses, arbitrary orientations, and complex backgrounds, the RHP dataset is more challenging than the STB dataset.
	}
	\label{fig:dataset}
	  \vspace{-1em}
\end{figure}

\begin{table*}[htb!]
\caption{Quantitative comparison of the generated hand images using {\em CycleGAN}, {\em Pix2pix}, {\em $\text{PG}^2$}, {\em Pose-GAN}, {\em PATN} and our proposed {\em MM-Hand} on the two benchmark datasets STB and RHP. We adopt the metric of SSIM, IS, mask-SSIM, mask-IS and PCKb to quantitatively evaluate the quality of the generated hand images. IS and Mask-IS are not reliable due to the lack of hand-related classes in the ImageNet dataset, which the {\em Inception-v3} model is pretrained on. Similarly, SSIM's scores can be erratic due to the background noise. We use (masked) SSIM and (masked) IS here because they are widely adopted as evaluation metrics in pose guided person generation. PCKb is the most indicative metric which takes both visual realism and pose consistency into account. For all the metrics here, the higher value is always the better.} 
\vspace{-0.5em}
\centering 
\scalebox{1.0}{
\begin{tabular}{c|ccccc||ccccc}
\hline
& \multicolumn{5}{c||}{$\mathcal{X}_{STB}$} &  \multicolumn{5}{c}{$\mathcal{X}_{RHP}$}\\
\hline
{} & SSIM & IS & mask-SSIM & mask-IS & \textbf{PCKb} & SSIM & IS & mask-SSIM & mask-IS & \textbf{PCKb} \\ 
\hline
CycleGAN & 0.002 & 1.52 & 0.611 & 2.49  & 0.07 & 0.008 &  2.08  & 0.816 & \textbf{2.98} & 0.015\\ 
Pix2pix & 0.027  & 2.24 & 0.625  & \textbf{2.632} & 0.527  & 0.010  & 2.67  & 0.816 & 2.85  & 0.119 \\ 
$\text{PG}^2$ & 0.026  & 2.33 & 0.638  & 2.224  & 0.686  & 0.021  & 2.236  & 0.822  & 2.762  & 0.250 \\ 
Pose-GAN & 0.02 & 1.01 & 0.610  & 1.495 & 0.05 & 0.014 & 1.03  & 0.808   & 2.012 & 0.014  \\
PATN & 0.014  & \textbf{2.371}  & 0.656  & 2.276 & 0.564  & 0.054  & 2.348 & 0.830  & 2.532 & 0.248  \\ 
\hline
MM-Hand (Ours) & \textbf{0.115} & 2.187 & \textbf{0.677} & 2.53 & \textbf{0.688} & \textbf{0.078} & \textbf{2.376} & \textbf{0.844} & 2.747 & \textbf{0.619} \\ 
\hline
\end{tabular}}
\label{table:quant-comparison} 
\end{table*}

\begin{table*}[htb!]
\caption{The 3D hand pose estimation performance using $\mathcal{M}_{Hand3D}$ and $\mathcal{M}_{3D-HPM}$ on $\mathcal{X}_{STB}$ and $\mathcal{X}_{RHP}$, augmented by images generated by different methods under different portion $\alpha$ of the training set $\mathcal{X}_{tr}$ as the reduced training set. \textit{None} means no generative model is employed, thus not using data augmentation. We adopt the metric of mean EPE to evaluate the estimation performance. Lower EPE indicates better performance.} 
\vspace{-0.5em}
\centering 
\scalebox{1.0}{
\begin{tabular}{c|c|ccccc||ccccc}
\hline
& & \multicolumn{5}{c||}{$\mathcal{X}_{STB}$} &  \multicolumn{5}{c}{$\mathcal{X}_{RHP}$}\\
\hline
{} & {} & 0.2 & 0.4 & 0.6 & 0.8 & 1.0 & 0.2 & 0.4 & 0.6 & 0.8 & 1.0 \\ 
\hline
\multirow{7}{*}{$\mathcal{M}_{Hand3D}$} & None & 60.28 & 48.84 & 30.23 & 18.74 & 9.81  & 90.12 & 60.27 & 36.58 & 25.29 & 20.52\\ 
& CycleGAN & 80.27 & 82.57 & 75.39 & 72.56 & 9.81 & 90.29 & 78.29 & 60.29 & 40.29 & 20.52\\ 
& Pix2pix & 72.57 & 71.27 & 54.13 & 50.25 & 9.81 & 82.39 & 76.48 & 62.16 & 80.29 & 20.52\\ 
& $\text{PG}^2$ & 74.27 & 68.22 & 58.23 & 52.28 & 9.81 & 85.29 & 72.83 & 64.49 & 40.25 & 20.52\\ 
& PATN & 70.28 & 68.23 & 50.37 & 40.57 & 9.81 & 84.25 & 74.49 & 84.35 & 60.25 & 20.52 \\ 
& Pose-GAN & 72.58 & 69.27 & 52.85 & 39.56 & 9.81 & 94.59 & 84.38 & 67.83 & 45.59 & 20.52 \\
\cline{2-12} & MM-Hand (Ours) & \textbf{52.39} & \textbf{32.37} & \textbf{27.49} & \textbf{16.48} & 9.81 & \textbf{80.29} & \textbf{54.38} & \textbf{38.49} & \textbf{24.38} & 20.52\\ 
\hline
\hline
\multirow{7}{*}{$\mathcal{M}_{3D-HPM}$} & None & 64.16 & 48.91 & 33.00 & 35.51 & 15.71 & \textbf{52.21} & 50.38 & 47.36 & 45.43 & 35.86 \\ 
& CycleGAN & 111.75 & 54.55 & 51.71 & 47.02 & 15.71 & 66.63 & 59.63 & 57.59 & 61.67 &  35.86\\ 
& Pix2pix & 99.59 & 46.71 & 47.83 & 46.91 & 15.71 & 65.73 & 64.56 & 62.31 & 55.07 &  35.86\\ 
& $\text{PG}^2$ & 91.03 & 47.00 & 47.20 & 46.78 & 15.71 & 61.05 & 58.95 & 57.59 & 56.72 &  35.86\\ 
& PATN & 99.66 & 46.95 & 47.84 & 40.18 & 15.71 & 56.11 & 50.26 & 50.64 & 51.92 &  35.86\\
& Pose-GAN & 102.70 & 46.65 & 48.03 & 47.02 & 15.71 & 60.54 & 57.44 & 53.40 & 52.83 &  35.86\\
\cline{2-12} & MM-Hand (Ours) & \textbf{41.79} & \textbf{20.24} & \textbf{16.79} & \textbf{16.15} & 15.71 & 52.47 & \textbf{42.22} & \textbf{41.63} & \textbf{40.49} &35.86  \\ 
\hline
\end{tabular}}
\label{table:reduced-train} 
\end{table*}

\begin{table}[htb!]
\caption{Ablation study of MM-Hand using $3$D pose estimation on the RHP and STB datasets. $\text{AUC}_{20-50}$ is the area under the PCK curve between 20mm and 50 mm. The higher $\text{AUC}_{20-50}$ is the better.}
    \setlength{\tabcolsep}{0pt}
    \begin{tabular}{c|lc}
    \hline
    & \multicolumn{1}{c}{Ablation Approaches} & \multicolumn{1}{l}{$\text{AUC}_{20-50}$} \\ 
\hline
    \multirow{5}{*}{$\mathcal{X}_{RHP}\ $} & \ {a)} base + $\bm{c}_{p_s},\bm{c}_{p_t}$ & 0.904\\
    & \ {b)} base + $\bm{c}_{p_s},\bm{c}_{p_t}$ + $\bm{M}_n$ & 0.908\\
    & \ {c)} base + $\bm{c}_{p_s},\bm{c}_{p_t}$ + $\bm{d}_{p_s},\bm{d}_{p_t}$ + $\bm{M}_n$ & 0.912\\
    & \ {d)} base + $\bm{c}_{p_s},\bm{c}_{p_t}$ + $\bm{d}_{p_s},\bm{d}_{p_t}$ + $\bm{M}_n$ + GCT & 0.915\\
    & \ {e)} base + $\bm{c}_{p_s},\bm{c}_{p_t}$ + $\bm{d}_{p_s},\bm{d}_{p_t}$ + $\bm{M}_n$ + GCT + INNM & 0.929\\
\hline
\hline
    \multirow{5}{*}{$\mathcal{X}_{STB}$} & \ {a)} base + $\bm{c}_{p_s},\bm{c}_{p_t}$ & 0.984\\
    & \ {b)} base + $\bm{c}_{p_s},\bm{c}_{p_t}$ + $\bm{M}_n$ & 0.988\\
    & \ {c)} base + $\bm{c}_{p_s},\bm{c}_{p_t}$ + $\bm{d}_{p_s},\bm{d}_{p_t}$ + $\bm{M}_n$ & 0.994\\
    & \ {d)} base + $\bm{c}_{p_s},\bm{c}_{p_t}$ + $\bm{d}_{p_s},\bm{d}_{p_t}$ + $\bm{M}_n$ + GCT & 0.997\\
    & \ {e)} base + $\bm{c}_{p_s},\bm{c}_{p_t}$ + $\bm{d}_{p_s},\bm{d}_{p_t}$ + $\bm{M}_n$ + GCT + INNM & 0.999\\
    \hline
    \end{tabular}
    \vspace{-1em}
    \label{table:ablation}
\end{table}

\subsection{Experimental Settings}

\paragraph{\textbf{Baselines Approaches}} 

We select {\em CycleGAN} ~\cite{zhu2017unpaired}, {\em Pix2pix} ~\cite{isola2017image}, {\em $\text{PG}^2$}~\cite{ma2017pose}, {\em Pose-GAN} ~\cite{siarohin2018deformable}, {\em PATN} ~\cite{zhu2019progressive} as the baseline methods to be compared with our proposed {\em MM-hand}. 

{\em CycleGAN} learns unpaired image-to-image translation, by enforcing cycle consistency to push the source domain of 2D pose maps to be consistent with the target domain of realistic hand images. 
{\em Pix2pix} learns the translation from the 2D hand pose label maps to a real hand image. 
{\em $\text{PG}^2$} is a two-stage coarse-to-fine network that generates a person image under target pose from a source person image. 
{\em Pose-GAN} introduces deformable skip connections in the generator. 
{\em PATN} generates person images under the guidance of the target pose progressively via attention mechanism. 

\paragraph{\textbf{3D Hand Pose Estimators}} Unfortunately, several state-of-the-art estimators for further comparison ~\cite{zimmermann2017learning,cai2018weakly,mueller2018ganerated,yang2019disentangling,yang2019aligning,baek2019pushing,ge20193d,boukhayma20193d} have not released their training code. We adopt two hand pose estimators, Hand3D~\cite{zimmermann2017learning} and 3D-HPM (proposed), to assess the quality of our generated hand training images through the improvement of the model performance. 
We propose 3D-HPM\footnote{3D-HPM works by first passing the input through a series of convolution layers grouped in 6 stages. Next, it passes the encoded information through a another series of convolution layers, which are regularized by the ground truth 2D coordinates. Lastly, 3D-HPM uses a fully-connected layer to predict the $z$ information.} to evaluate 3D coordinates from a single monocular RGB image.
Although not as good as the state-of-the-art estimators, Hand3D and 3D-HPM have achieved EPE of 9.81 and 15.71 on the STB dataset, which are close to the reported state-of-the-art results.

\paragraph{\textbf{Datasets}} We select two benchmark datasets for performance evaluation including the \textit{Stereo Tracking Benchmark} (STB)~\cite{zhang20163d}, and the \textit{Render Hand Pose} (RHP)~\cite{zb2017hand}.
The RHP dataset contains $41,258$ training and $2,728$ testing hand samples captured from $20$ subjects performing $39$ actions. 
Each sample consists of an RGB image, a depth map, and the segmentation masks for the background, person, and each finger. 
Each hand is annotated with its $21$ keypoints in both 2D image coordinates and 3D world coordinate positions. 
The RHP dataset is split into a validation set (R-val) and a training set (R-train).
This dataset is challenging due to the large variations in viewpoints and the low image resolution. 
The STB dataset has $18,000$ hand images. 
It is split into two subsets: the stereo subset (STB-BB) and the color-depth subset (STB-SK).

\paragraph{\textbf{Evaluation Metrics}}

To evaluate the visual realism and pose consistency, we propose PCKb and adopt SSIM and IS with their masked version for evaluation.

The PCKb is the percentage of correct keypoints within two-thirds of the average length of the bones in a hand image. PCKb is averaged across the given hand images for evaluation. Mask-SSIM and mask-IS are evaluated on the hand image with a masked background.

We further adopt two metrics for evaluating the accuracy of estimated hand poses: 
the average End-Point-Error (EPE), and the Area Under the Curve (AUC) on the Percentage of Correct Keypoints (PCK). 
The performance metrics are computed in pixels (px) on the hand images and millimeters (mm) in 3D world coordinate, respectively. 
The performance of 3D hand joint prediction is measured by the PCK curves averaged over all $21$ keypoints.

\vspace{-1em}

\subsection{Experimental Results}
\begin{figure*}[t]
    \setlength{\tabcolsep}{0.2em}
    \renewcommand{\arraystretch}{0}
	\begin{tabular}{c@{}c@{}c@{}c@{}c@{}c@{}c@{}c}
		\includegraphics[width=0.125\textwidth]{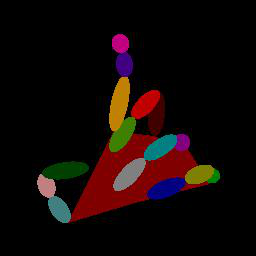} &
		\includegraphics[width=0.125\textwidth]{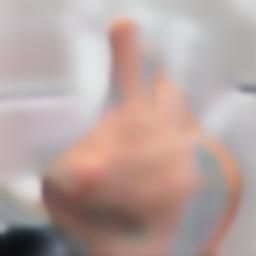} &
		\includegraphics[width=0.125\textwidth]{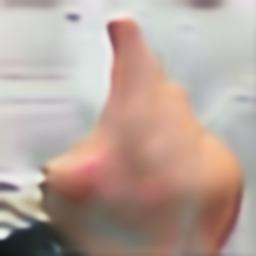} &
		\includegraphics[width=0.125\textwidth]{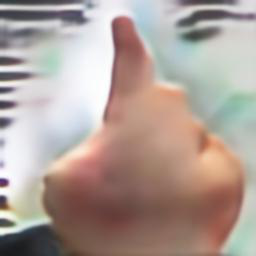} &
		\includegraphics[width=0.125\textwidth]{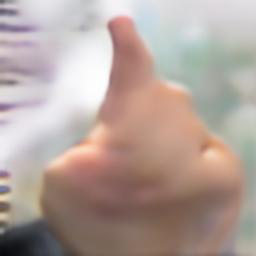} &
		\includegraphics[width=0.125\textwidth]{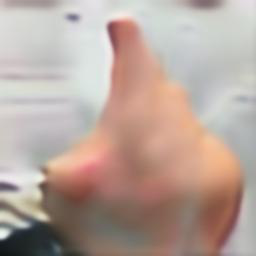} &
		\includegraphics[width=0.125\textwidth]{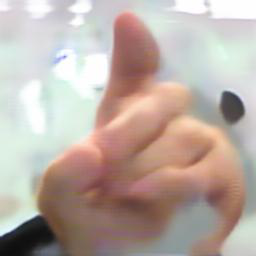} &
		\includegraphics[width=0.125\textwidth]{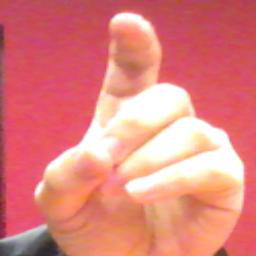} \\
		\label{fig:stb}
	\end{tabular}
	\hfill
	\begin{tabular}{c@{}c@{}c@{}c@{}c@{}c@{}c@{}c}
		\includegraphics[width=0.125\textwidth]{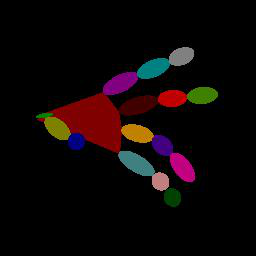} &
		\includegraphics[width=0.125\textwidth]{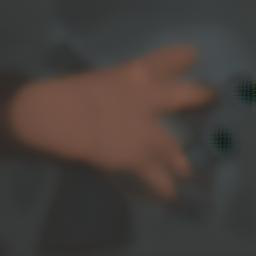} &
		\includegraphics[width=0.125\textwidth]{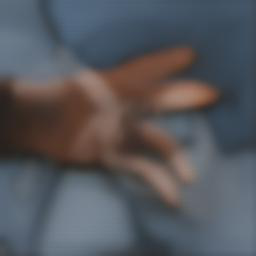} &
		\includegraphics[width=0.125\textwidth]{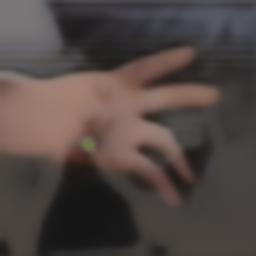} &
		\includegraphics[width=0.125\textwidth]{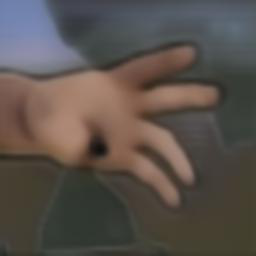} &
		\includegraphics[width=0.125\textwidth]{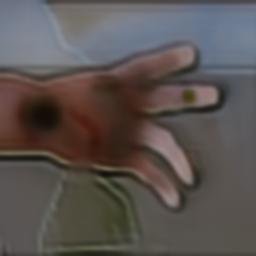} &
		\includegraphics[width=0.125\textwidth]{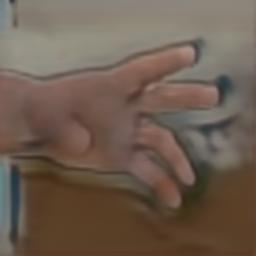} &
		\includegraphics[width=0.125\textwidth]{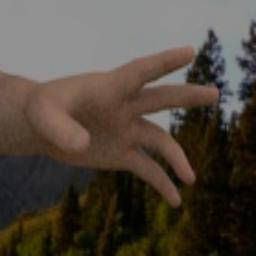} \\
		$\bm{p_t}$ & \textbf{CycleGAN} & \textbf{Pix2pix} & $\textbf{PG}^2$ & \textbf{Pose-GAN} & \textbf{PATN} & \textbf{MM-Hand} & \textbf{Ground-truth}
		\label{fig:rhp}
	\end{tabular}
	\vspace{-1em}
	\caption{Qualitative comparison of the synthesized hand images using {\em CycleGAN}, {\em Pix2pix}, {\em $\text{PG}^2$}, {\em Pose-GAN}, {\em PATN} and our proposed {\em MM-Hand} on the two benchmark datasets STB and RHP. \textbf{From top to bottom:} the STB dataset and the RHP dataset.}
	\label{fig:qual-comparison}
\end{figure*}

\subsubsection{Qualitative and Quantitative Comparison of the Synthesized Hand Images}

The source domain is 2D hand pose maps derived from 3D hand pose by projection, and the target domain is realistic hand images.
The source domain is represented by 2D hand pose label maps. Figure~\ref{fig:qual-comparison} and Table~\ref{table:quant-comparison} show the visual quality and the quantitative comparison of the hand images generated by {\em CycleGAN}, {\em Pix2pix}, {\em $\text{PG}^2$}, {\em Pose-GAN}, {\em PATN}, and our proposed~{\em MM-hand}. Our proposed {\em MM-Hand} beats the baseline approaches with better visual quality and higher values in most of the metrics on both STB and RHP. Note that PCKb is the most indicative metric that measures both visual realism and pose consistency. {\em MM-Hand} is able to beat other approaches with a large margin on PCKb.

\subsubsection{Boosting the Performance of 3D Hand Pose Estimators when Training Data is Reduced}

The hand image generation method can generate realistic hand images for improving the 3D hand pose estimator learning, especially when the original training data is insufficient.
In the training set $\mathcal{X}_{tr}$ of the 3D hand pose dataset, we randomly select a portion $\alpha$ of $\mathcal{X}_{tr}$ as the reduced training set, denoted as $\mathcal{X}^R_\alpha$.
The rest of the data is denoted as $\mathcal{X}^R_{1-\alpha}$ (\ie $\mathcal{X}^R_{1-\alpha} \cup \mathcal{X}^R_\alpha = \mathcal{X}_{tr}$). 

Using the poses ($\bm{p}_t$s) in $\mathcal{X}^R_{1-\alpha}$ and the images with poses ($\bm{I}_{p_s}$s and $\bm{p}_s$s) in $\mathcal{X}^R_\alpha$, we build $\overline{\mathcal{X}}^R_{1-\alpha}$ by replacing each image $\bm{I}_{p_t}$ in $\mathcal{X}^R_{1-\alpha}$ with $\bm{\Tilde{I}}_{p_t}$. Given $\bm{p}_t$, $\bm{\Tilde{I}}_{p_t}$ is generated from the best matched source hand image $\bm{I}_{p_s}$ in $\mathcal{X}^R_\alpha$, whose pose $\bm{p}_s$ is closest to $\bm{p}_t$ in pose distance. Specifically,
\begin{align}
    \begin{split}
        \forall (\bm{I}_{p_t}, \bm{p}_t) \in \mathcal{X}^R_{1-\alpha},\ 
        & \exists (\bm{I}_{p_s^*}, \bm{p}_s^*) \in \mathcal{X}^R_\alpha,\\
        \text{s.t.}\ \bm{p}_s^*=\argmin_{\bm{p}_s} d(\bm{p}_s, \bm{p}_t),\ 
        & <\bm{I}_{p_s^*}, \bm{p}_s^*, \bm{p}_t> \rightarrow \bm{\Tilde{I}}_{p_t}.
    \end{split}
\end{align}

Consequently, an augmented training set $\mathcal{X}^A_\alpha$ is formed from $\mathcal{X}^R_\alpha$ and $\overline{\mathcal{X}}^R_{1-\alpha}$ (\ie $\mathcal{X}^R_\alpha \cup \overline{\mathcal{X}}^R_{1-\alpha} = \mathcal{X}^A_\alpha$). We use $\mathcal{X}^A_\alpha$ instead of  $\mathcal{X}_{tr}$ to train the two 3D hand pose estimators (Hand3D and 3D-HPM) respectively. All the numbers in Table \ref{table:reduced-train} are obtained by training the 3D hand pose estimators on the augmented training set $\mathcal{X}_\alpha^A$ and evaluating the estimators on the testing set $\mathcal{X}_{te}$.

Inspired by Zhu \etal~\cite{zhu2019progressive}, this experiment is designed to assess the quality of our generated hand images, which would boost the performance of 3D hand pose estimation, as augmented data. 

Table \ref{table:reduced-train} shows that augmenting the training data of STB and RHP using hand images generated by our {\em MM-Hand} can achieve performance gain across different values of $\alpha$. Moreover, the performance gain is more significant if the performance gap is relatively large.

We further compare our approach to {\em $\text{PG}^2$}, {\em Pose-GAN}, and {\em PATN} under the same experimental setting for 3D hand pose estimation. 
%
Table~\ref{table:reduced-train} shows that our approach achieves consistent improvements over all previous works for different $\alpha$s, suggesting the proposed method can generate more realistic and pose-preserving hand images for the 3D hand pose estimation task. 

\begin{figure*}[htb!]
\begin{tabular}{cc}
\centering 
{\includegraphics[width=0.485\textwidth]{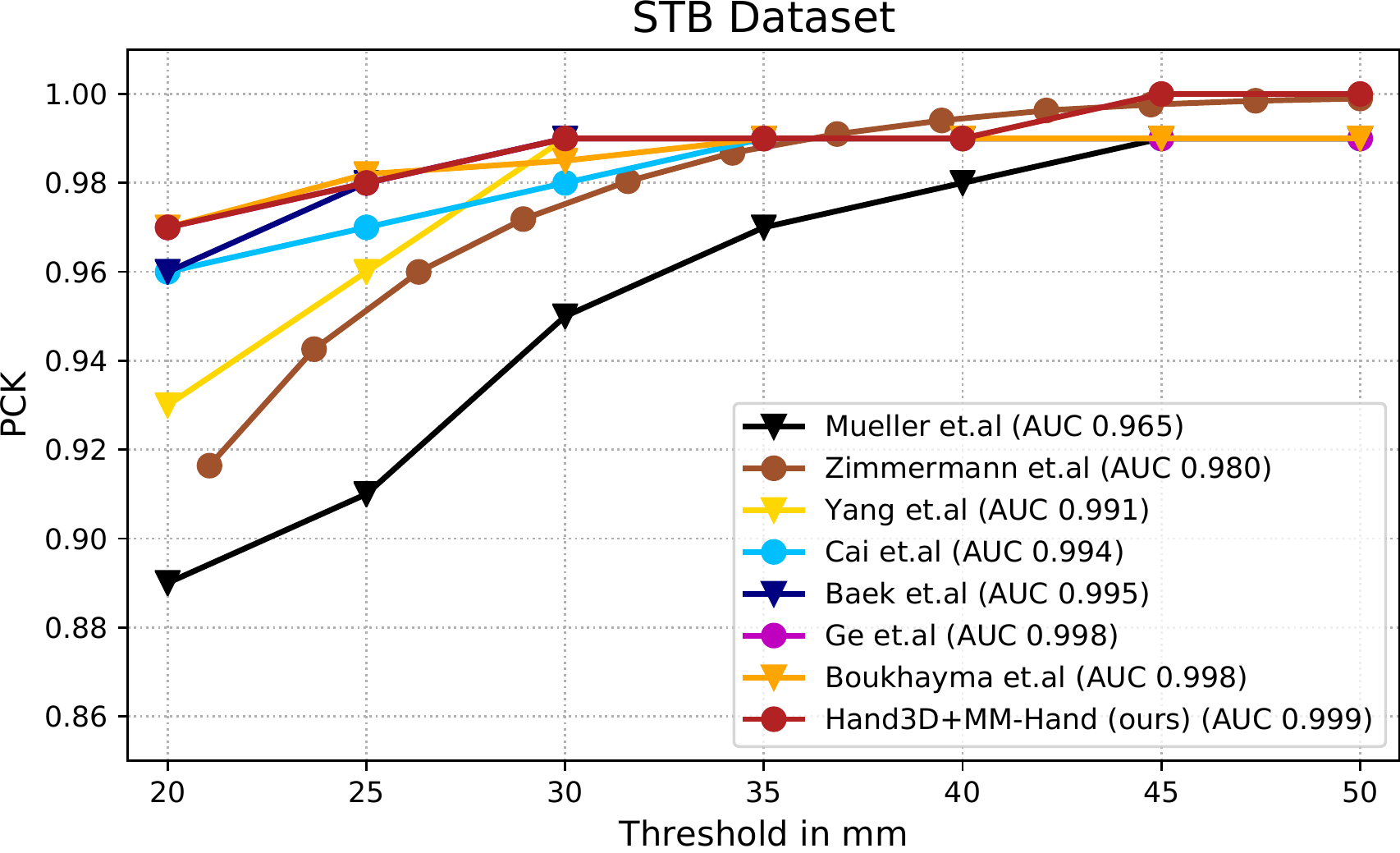}}\label{fig:3DSTB}&
{\includegraphics[width=0.485\textwidth]{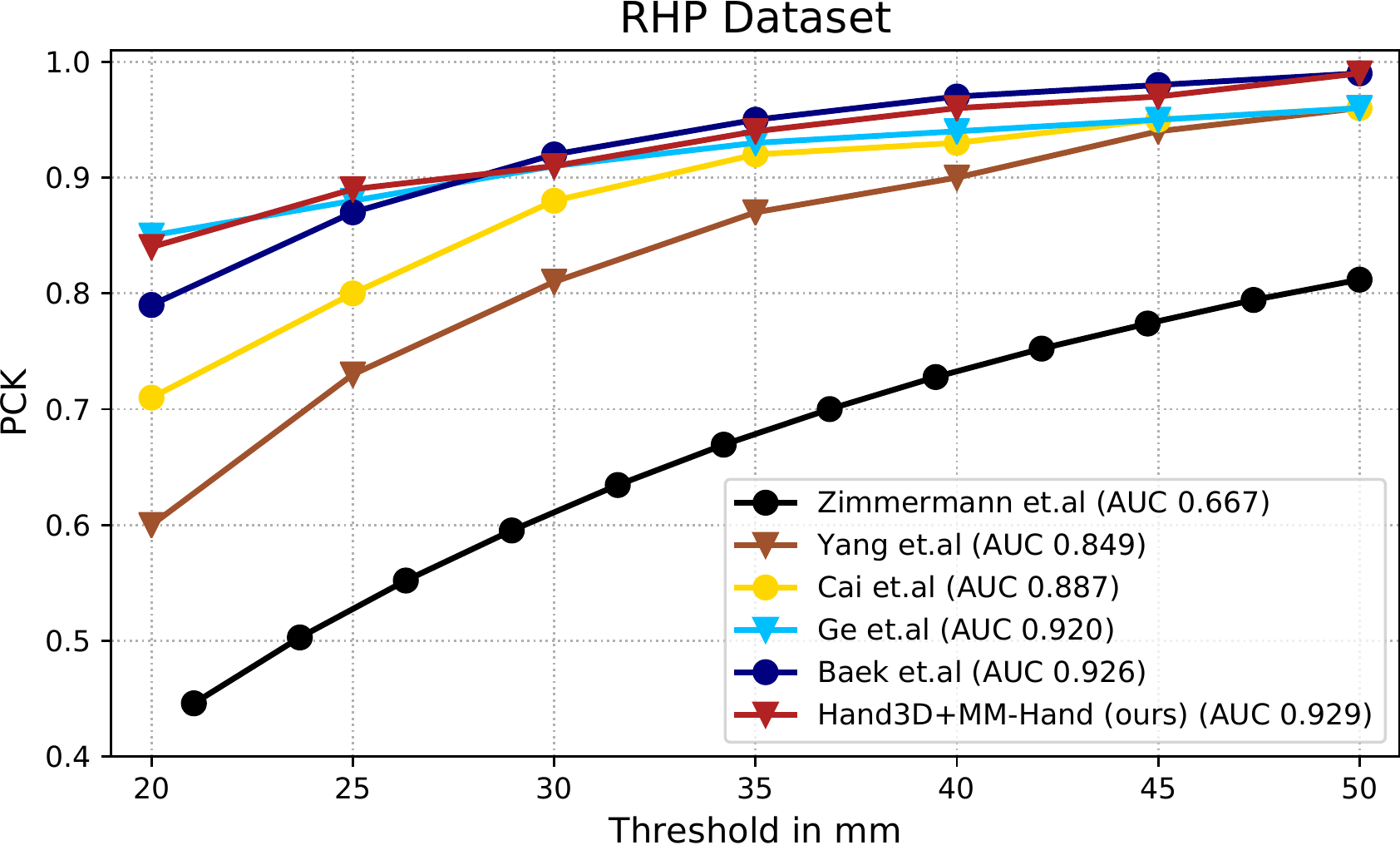}\label{fig:3DRHP}}\\
\end{tabular}
\vspace{-1em}
\caption{Our comparisons with state-of-the-art approaches on (a) the STB dataset in 3D hand pose estimation task; (b) the RHP dataset in 3D hand pose estimation task. We choose all the state-of-the-art approaches proposed after 2017.}
\label{fig:comp-SOTA-AUC}
\end{figure*}

\begin{figure*}[htb!]
\setlength{\tabcolsep}{0em}
\begin{tabular}{cccccccc}
		\includegraphics[width=0.12\textwidth]{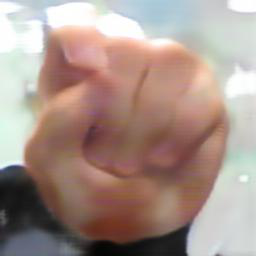}
		\includegraphics[width=0.12\textwidth]{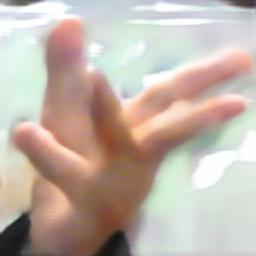}
		\includegraphics[width=0.12\textwidth]{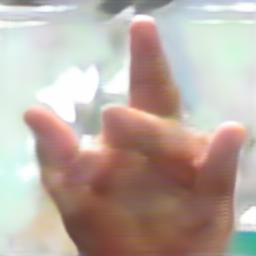}
		\includegraphics[width=0.12\textwidth]{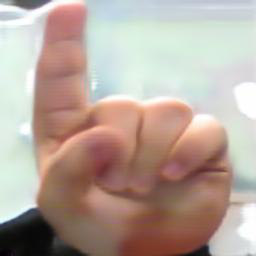}
		\includegraphics[width=0.12\textwidth]{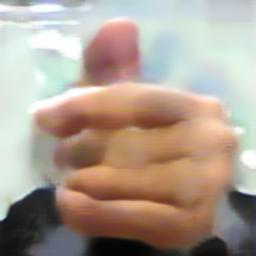}
		\includegraphics[width=0.12\textwidth]{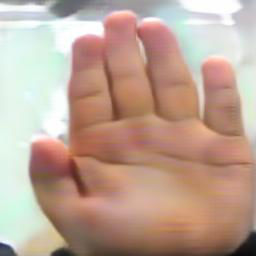}
		\includegraphics[width=0.12\textwidth]{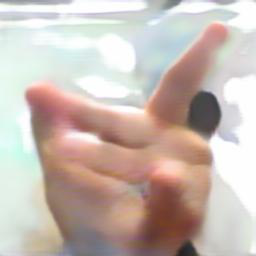}
		\includegraphics[width=0.12\textwidth]{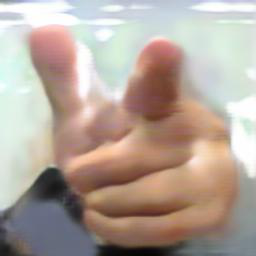} \\
		\includegraphics[width=0.12\textwidth]{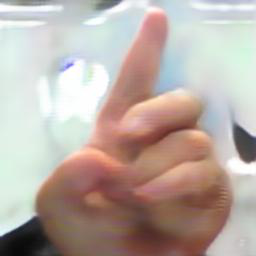}
		\includegraphics[width=0.12\textwidth]{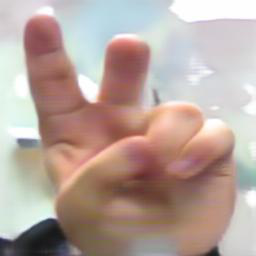}
		\includegraphics[width=0.12\textwidth]{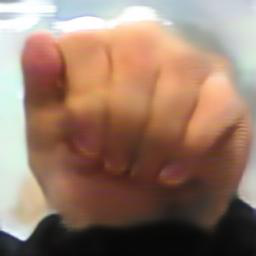}
		\includegraphics[width=0.12\textwidth]{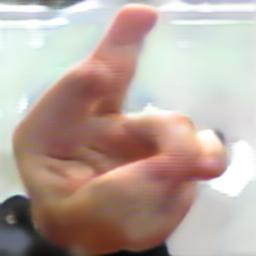}
		\includegraphics[width=0.12\textwidth]{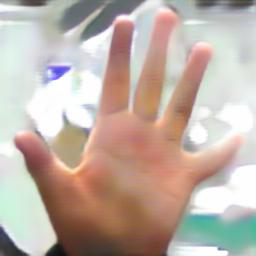}
		\includegraphics[width=0.12\textwidth]{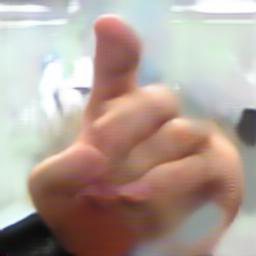}
		\includegraphics[width=0.12\textwidth]{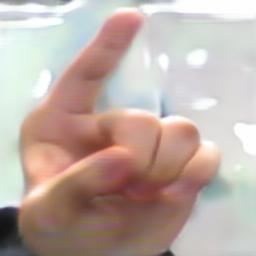}
		\includegraphics[width=0.12\textwidth]{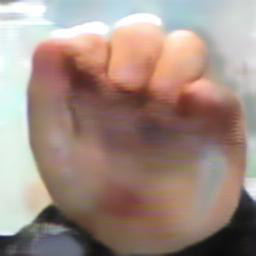} \\
		\label{fig:stb}
	\end{tabular}
	\vspace{-2.0em}
	\caption{We randomly pick $16$ hand images generated by our proposed {\em MM-Hand} trained on STB. Despite some missing texture details, the generated hand images look realistic and are consistent with the target poses $p_t$s obtained from Blender.}
	\label{fig:visualization}
\end{figure*}

\subsubsection{Ablation Study of MM-Hand}
To verify the effectiveness of the proposed 3D pose embeddings and the geometry-based curriculum learning, we incrementally evaluate them on the STB dataset. 
We choose ResNetGenerator as the baseline method. 
Five variants are then constructed on top of the baseline: 
\begin{itemize}[leftmargin=*]
    \item[] \textbf{a)} ResNetGenerator (base) + contour map ($\bm{c}_{p_s},\bm{c}_{p_t}$);
    \item[] \textbf{b)} ResNetGenerator (base) + contour map ($\bm{c}_{p_s},\bm{c}_{p_t}$) + attention mask ($\bm{M}_n$);
    \item[] \textbf{c)} ResNetGenerator (base) + contour map ($\bm{c}_{p_s},\bm{c}_{p_t}$) + depth map ($\bm{d}_{p_s},\bm{d}_{p_t}$) + attention mask ($\bm{M}_n$);
    \item[] \textbf{d)} ResNetGenerator (base) + contour map ($\bm{c}_{p_s},\bm{c}_{p_t}$) + depth map ($\bm{d}_{p_s},\bm{d}_{p_t}$) + attention mask ($\bm{M}_n$) + geometry-based curriculum training (GCT);
    \item[] \textbf{e)} ResNetGenerator (base) + contour map ($\bm{c}_{p_s},\bm{c}_{p_t}$) + depth map ($\bm{d}_{p_s},\bm{d}_{p_t}$) + attention mask ($\bm{M}_n$) + geometry-based curriculum training (GCT) + inference with nearest neighbor match (INNM).
\end{itemize}

Table ~\ref{table:ablation} presents the ablation study results, from which we can draw a conclusion that incrementally adding pose embedding (contour map + depth map), attention mask, geometry-based curriculum training, and inference nearest neighbor match consistently improve the 3D hand pose estimation performance.

\subsubsection{Surpassing State-of-the-Art 3D Hand Pose Estimators under Standard Setting}
We compare with the current state-of-the-art 3D hand pose estimators, including Z\&B~\cite{zimmermann2017learning}, Cai~\etal~\cite{cai2018weakly}, Mueller~\etal~\cite{mueller2018ganerated}, Yang~\etal~\cite{yang2019disentangling}, Baek~\etal~\cite{baek2019pushing}, Ge~\etal~\cite{ge20193d}, Boukhayma~\etal~\cite{boukhayma20193d}, Yang~\etal~\cite{yang2019aligning}. 
%
We report our results by training the Hand3D model with additional hand images generated by our proposed {\em MM-Hand} for augmentation. 
%
%
Unfortunately, all the state-of-the-art methods have not (or partially) released their code at the time of writing.
Hence, we select Hand3D in our experiment, which provides both training and evaluation code.
Figure~\ref{fig:comp-SOTA-AUC} and Table~\ref{table:comp-SOTA-EPE} demonstrate that training Hand3D with additional hand images generated by {\em MM-Hand} achieves the best performance on STB and RHP in AUC of the PCK curve. 
We further show some examples of our generated hand images in Figure~\ref{fig:visualization}. 
These images are generated by {\em MM-Hand} trained on the STB dataset. 
The target poses $\bm{p_t}$s are generated using an augmented reality (AR) simulator, \ie Blender\footnote{https://www.blender.org}.

\begin{table}[htb!]
\centering
\caption{3D hand pose estimation results on the RHP and STB datasets. $\text{AUC}_{20-50}$ is the area under the PCK  curve between 20mm and 50 mm. The higher $\text{AUC}_{20-50}$ is the better.}
    \begin{tabular}{c|cc}
    \hline
    & Approaches & {$\text{AUC}_{20-50}$} \\ 
\hline
    \multirow{6}{*}{$\mathcal{X}_{RHP}$} & {Z\&B ~\cite{zb2017hand}} & 0.667 \\
    & {Yang \etal~\cite{yang2019disentangling}} & 0.849 \\
    & {Cai \etal~\cite{cai2018weakly}} & 0.887  \\
    & {Ge \etal~\cite{ge20193d}} & 0.92\\
    & {Baek \etal~\cite{baek2019pushing}} & 0.926 \\
    & Hand3D + MM-Hand (Ours) & \textbf{0.929}\\
\hline
\hline
    \multirow{6}{*}{$\mathcal{X}_{STB}$} & {Z\&B ~\cite{zb2017hand}} & 0.980\\
    & {Yang \etal~\cite{yang2019disentangling}} & 0.991\\
    & {Cai \etal~\cite{cai2018weakly}} & 0.994 \\
    & {Baek \etal~\cite{baek2019pushing}} & 0.995\\
    & {Ge \etal~\cite{ge20193d}} & 0.998 \\
    & Hand3D + MM-Hand (Ours) & \textbf{0.999}\\
    \hline
    \end{tabular}

    \label{table:comp-SOTA-EPE}
\end{table}

\section{Conclusion}
Due to the inherent depth ambiguity, building a sizeable real-world hand dataset with accurate 3D annotations is one major challenge of 3D hand pose estimation. 
We propose a 3D-aware multi-model guided hand generative network ({\em MM-Hand}), and a novel geometry-based training and inference strategy to generate hand images under the guidance of 3D hand poses.
With the help of an external dataset with paired depth maps and 3D hand poses, we train the depth map generator to synthesize depth maps based on any given 3D poses. 
Our proposed {\em MM-Hand} can generate realistic, diverse, and pose preserving hand images based on any given 3D poses and synthetic depth maps. 
Qualitative and quantitative results show that the hand images generated by {\em MM-Hand} resemble the ground truth hand images in both appearance and pose. 
Moreover, the hand images augmented by our proposed {\em MM-Hand} can consistently improve the 3D hand pose estimation results under different reduction portion.
We further verify the effectiveness of all proposed strategies using ablation study.
Additionally, we evaluate the quality of our generated hand images using poses obtained using an AR simulator (Blender) on two benchmark datasets. 
Our experimental results show that training 3D hand pose estimator with our augmented data can outperform the state-of-art methods on both STB and RHP datasets.

\begin{acks}
The work was mainly done during Z. Wu and D. Hoang's internships at Tencent America. The research of Y.-Y. Lin was supported by Ministry of Science and Technology under grants 107-2628-E-009-007-MY3, 109-2634-F-007-013, and 109-2221-E-009-113-MY3. The research of Z. Wang was partially supported by US Army Research Office Young Investigator Award W911NF2010240.
\end{acks}


\end{document}